# Understanding Soft Tissue Behavior for Application to Microlaparoscopic Surface Scan

Mustafa Suphi Erden, Benoît Rosa, Jérôme Szewczyk, and Guillaume Morel, *Members, IEEE*

*Abstract*—This paper presents an approach for understanding the soft tissue behavior in surface contact with a probe scanning the tissue. The application domain is confocal microlaparoscopy, mostly used for imaging the outer surface of the organs in the abdominal cavity. The probe is swept over the tissue to collect sequential images to obtain a large field of view with mosaicking. The problem we address is that the tissue also moves with the probe due to its softness; therefore the resulting mosaic is not in the same shape and dimension as traversed by the probe. Our approach is inspired by the finger slip studies and adapts the idea of load-slip phenomenon that explains the movement of the soft part of the finger when dragged on a hard surface. We propose the concept of loading-distance and perform measurements on beef liver and chicken breast tissues. We propose a protocol to determine the loading-distance prior to an automated scan and introduce an approach to compensate the tissue movement in raster scans. Our implementation and experiments show that we can have an image-mosaic of the tissue surface in a desired rectangular shape with this approach.

*Index Terms*—Laparoscopy, imaging, medical robotics, mosaicking, scan, soft tissue.

## I. INTRODUCTION

THIS paper presents a first attempt to understand the soft tissue behavior at the contact surface while being scanned with a rigid probe for confocal microscopy. Confocal microlaparoscopy is a promising approach in minimally invasive surgery for replacing conventional biopsies that involve physical tissue sampling. There have been different designs for confocal micro imaging of living tissues in microlaparoscopy [1, 2] and closely related microendoscopy [3, 4, 5, 6]. The images obtained by such techniques typically cover an area of 240×200 µm$^2$. The smallness of the image size is because of the necessity of a fine resolution and because the lenses and the fiber optic cable are minimized for minimal invasiveness. Such an image is usually not enough for a conclusive diagnostics, which typically requires covering an area of a few mm$^2$.

A solution to obtain images with larger field of view is to scan the region of interest and merge the collected images by mosaicking algorithms [7, 8]. The research in [9] demonstrates the applicability of this approach to image-mosaicking in vivo on human patients by manually passing the miniprobe over the region of interest and by using the mosaicking algorithm in [10]. Image-mosaicking with confocal microscopy is performed also on human-hand skin by again manually dragging a MEMS based scanner design [11] and using the mosaicking algorithm in [12]. Although these studies show the feasibility of image-mosaicking with confocal microscopy, they do not propose an automated tissue scan.

A typical scan for image-mosaicking in laparoscopic operations requires a position precision of up to 50-100 µm for duration of typically one or two minutes. With manual sweeping it is difficult to obtain this precision. Assistive handheld instruments are presented for micro positioning, for intraocular laser surgery [13] and for confocal laser endomicroscopy [14], but they are large to be used in minimally invasive surgery and are not intended for long and continuous manipulation. The motorized surgical microscope presented in [15] enables the surgeon to control the movement of the microscope by index finger movements with a remote controller. Although, this is a semi-automated system, it would be tedious for a surgeon to make the probe follow a proper scan path by finger movements.

A recent study by the latter three authors and colleagues demonstrates a design for automated confocal microlaparoscopic scanner based on hydraulic balloon catheters actuation [16]. The system is mounted inside a 5 mm inner diameter tube and enables active control of the trajectory of the probe moving over and in contact with the soft tissue. An important issue with such a system is that the surface of the soft tissue deforms under the contact with the probe. Due to the deformation the trajectory of the probe with respect to the tissue surface deviates from the trajectory of the probe with respect to the global reference frame. Therefore, a correction action is needed to compensate for the tissue motion and deformation [17]. The question is how to design the trajectory of the probe with respect to the global reference frame in order to obtain the desired trajectory of the probe with respect to the tissue surface. In order to answer this question we need to know how the soft tissue behaves when it is subject to a friction force on the surface. This paper contributes to understanding such tissue behavior. We propose the concept of loading-distance inspired by the finger slip studies, present the results of soft-tissue experiments, and develop an approach for compensation of tissue deformation in raster scan.

Laser micro imaging technologies construct a single image either by illuminating selectively distinguished points of the region under the probe [18, 19], by selective reception of light from these points [20], or by both [11]. In all these

Manuscript received March 10, 2012.

This research was funded by OSEO (Maisons-Alfort, France) under ISI Project PERSEE (number I0911038W). The partners are Mauna Kea Technologies (project leader, Paris, France), Endocontrol (Grenoble, France), Institut Mutualiste Montsouris (Paris, France), Institut Gustave Roussy (Villejuif, France) and ISIR-UPMC (Paris, France).

M.S. Erden was with the Institut des Systèmes Intelligents et de Robotique (ISIR) at Université Pierre et Marie Curie (UPMC), Paris, France. He is now with the LASA Laboratory at École Polytechnique Fédérale de Lausanne (EPFL), CH 1015, Switzerland. (corresponding author: + 41 (0) 21 69 32 919 ; email: mustafasuphi.erden@gmail.com).

B. Rosa, J. Szewczyk, and G. Morel are with ISIR at UPMC, Paris, France (e-mail: {rosa, sz, morel}@isir.upmc.fr).

technologies the emission and reception of the laser light is through the probe. This paper is about moving the probe on the soft tissue in order to collect such constructed sequential single images.

In the literature there are various studies about modeling soft tissue behavior considering the reaction to force impacts [17]. Most of these studies aim at modeling the reaction forces against indentation [21] or pulling effect with solid instruments [22] and most of them are intended for simulation purposes in virtual environments [23]. The finite-element modeling is a common continuum-based approach for modeling the soft-tissue behavior [23, 24]. The mass-spring model is a discrete approach approximating the continuous tissue structure by a finite set of nodes and massless-springs [23, 25]. None of these studies dwell upon the behavior of soft tissue under frictional effects. The research in [26] concerning a colonoscopy simulator makes use of the load and slip phases like in our paper, in order to simulate the friction forces with *a priori* known friction coefficients. Implementation of such force related modeling of tissue behavior in a practical system would imply integration of a force sensor to precisely measure the friction forces and having *a priori* knowledge of the model parameters of the tissue. These are challenging with a minimally invasive instrument and with the varieties of soft tissue structures in practice. Therefore, we target at an easy and practical approach for compensation of tissue deformation, which does not require force measurements and *a priori* knowledge of model parameters.

Our research is inspired by the finger slip studies [27, 28, 29]. These studies point out to the load-slip phenomenon observed in the soft tissue of the finger under the impact of a frictional surface contact. We make use of this idea to explain the behavior of the soft tissue being scanned with the probe. We aim at understanding how the tissue reacts, by comparing the movement of the tissue with the movement of the probe. The parameter developed to understand the tissue behavior, the loading-distance, makes reference to the spatial distance that the probe covers before the slipping starts. A preliminary version of this paper is presented [30]. In the current paper we extend the loading-distance compensation to a full and compact raster scan. We integrate the loading-distance measurement and the compensation with the raster scan that they are automatically performed with one stroke. Besides, further soft tissue measurements and scans are performed to demonstrate the repeatability and reliability of the approach.

The approach developed in this paper might be used with a minimally invasive device like in [16]. The surgeon can manually position and stabilize the device on the tissue and give the command of scan in one go either by pressing a button or by other means. Then the system can automatically make the loading-distance measurement and generate a probe motion which compensates for the tissue deformation. In this way a tissue scan covering an area of around 1 mm$^2$ can be performed in less than one minute.

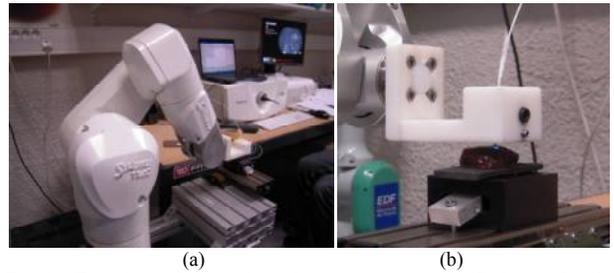

Fig. 1. Experimental setup for soft tissue scanning. (a) Left to right: the Stäubli-Robot, the computer controlling the robot, and the computer performing the image acquisition. (b) The soft tissue (beef liver) under the probe attached to the end effector of the robot.

## II. EXPERIMENTAL SETUP

In our experiments we use the Cellvizio system (Mauna Kea Technologies, Paris, France) [20] that performs confocal fluorescence imaging by recording light reception from selectively distinguished points. The system acquires images of size 240×200 μm$^2$ with 1.4 μm lateral and 10 μm axial resolutions at 12 frames/sec. The excitation wavelength is 488 nm. For proper illumination we apply Acriflavine on the tissue as the fluorescent agent. The system is equipped with the mosaicking algorithms presented in [10, 31]. We briefly review these algorithms and their precision in Sections III.A and III.C. The overall confocal-probe consists of a flexible bundle of optical fibers and an optic-head hosting the micro lenses, located at the tip. The outer diameter of the flexible bundle is 1.4 mm. The optic-head is a 12 mm long cylinder with an outer diameter 2.6 mm. In this paper we refer to the optic-head by the word "probe".

The experiments are performed with a six degrees-of-freedom robot (Stäubli TX40) on beef liver and chicken breast purchased from the supermarket (Fig. 1). With our control algorithm the deviation of the robot from a commanded trajectory might be up to around 50 μm, especially at the sharp corner turns. We name the trajectory of the probe with respect to the global reference frame as the *probe trajectory*. We measure the probe trajectory using the Cartesian space measurements of the robot and pass the position signals through a low pass filter in order to eliminate the measurement noise.

The trajectory of the probe with respect to the tissue surface is named as the *image trajectory*. The image trajectory deviates from the probe trajectory; because the tissue also moves under the impact of the movement of the probe. The image trajectory can be captured by adding successive translation vectors obtained by comparing successive images. The imaging algorithm in our Cellvizio system [31] performs this calculation and returns the position difference between the centers of sequential images throughout the scan.

Throughout our experiments we maintain an indentation depth of 350±50 μm. We first translate the probe vertically in steps of 100 μm distance until it gets into contact with the tissue. We determine the moment of contact by observing the images on the monitor. Then we further intrude the tissue with a distance 300 μm. In this way, each case we maintain the same indentation depth with an error margin 100 μm. In [23], it is demonstrated that the force vertically applied on a soft tissue is almost proportional to the indentation depth. This suggests that we can assume same pressure level each time we perform measurement on the same tissue. Before each experiment we hydrate the soft-tissue with isotonic saline against drying. Therefore, we can assume same hydration condition in the contact surface for all our measurements.

### III. QUALITY AND PRECISION OF MEASUREMENTS

In this paper there are three kinds of position measurements: 1) the robot measurements for the position of the probe with respect to the global reference frame, 2) the online measurements by the imaging algorithm for the position of the probe with respect to the scan surface, 3) the offline measurements by the imaging algorithm for the position of the probe with respect to the scan surface. In this paper we use the offline measurements as a reference for the performance of the robot and online measurements. Therefore, we explain first the offline measurement, then the robot measurement, and lastly the online measurment.

#### A. Precision of Offline Image Mosaicing

The offline mosaicking is the name for the main algorithm that constructs the mosaic by running an optimization after the scan. The measurement by this algorithm refers to the position of the center of each image merged to the overall mosaic. The optimization takes the result of the online measurement that will shortly be explained as an initial guess and refines it iteratively. The offline measurement has a high precision up to sub-microns [32, 10]. Here we perform mosaics with this algorithm on a paper where 0.33 mm distanced grid lines are laser-printed, in order to compare its performance with respect to a ground truth image.

In Fig. 2(a) we present the offline mosaic after a raster scan of an area $1.65 \times 1.65$ mm$^2$ on the paper. The original grid lines are shown in Fig. 2(b) with the mosaic superimposed within the red rectangle and with one millimeter distanced stripes of a ruler at the bottom. We see in Fig. 2(b) that the vertical and horizontal lines are reproduced almost identically as in the ground-truth image. In Fig. 2(a), both the successive images on a single raster line and the images on different raster lines are successfully merged without any gap. The distance between the grid lines is 0.33 mm as in the ground truth image. This is an indication that, first, we can trust the output mosaic of the offline algorithm for properly merging the successive images, and second, we can use the position output of the offline algorithm as a reference to compare the robot and online measurements.

#### B. Precision of the Robot

In order to quantify the precision of the robot measurements we perform circular scans of 1 mm diameter on the laser-printed paper and construct the offline mosaics as shown in Fig. 2(c). The distance between the first and last images is compared with the distance between the start and end positions of the robot end effector. The distance between the first and last images is calculated by applying the offline mosaicking algorithm. In total we performed 10 circle scans.

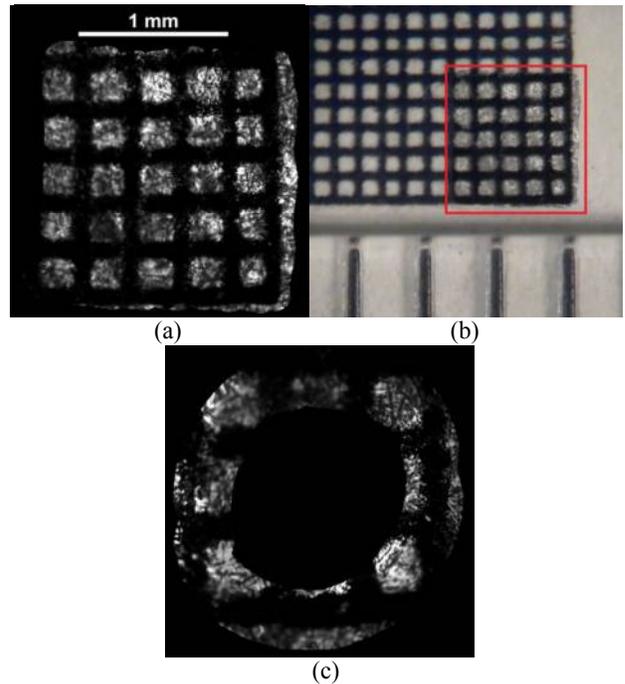

Fig. 2. (a) An offline image mosaic obtained by scanning the grid on the surface of a laser-printed paper. (b) The mosaic superimposed on the image of the laser-printed paper. (c) The offline mosaic with a circle scan of 1 mm diameter on the laser-printed paper.

Over the ten measurements the mean difference between the image and robot measurements is 3.53 μm with a standard deviation 3.75, a minimum 0.65 and a maximum 14.22. This result indicates that the maximum error of the robot measurement is far less than the single image size ($200 \times 240$ μm$^2$).

#### C. Precision of Online Distance Measurement

The imaging algorithm can perform online image mosaicking by a fast algorithm based on the estimation of the translation between two successive images using a 2D normalized cross correlation. The algorithm evaluates, in one pass, the correlation coefficient between the successive images for every possible translation and returns the translation value that results in the largest correlation. The position of each image is calculated by integrating the positions of the preceding images. The details and a clinical application of the algorithm are presented in [31]. Unlike the case with the offline mosaicking there is no global optimization to accommodate for the errors in individual position measurements. Therefore, the integration of the errors results in a drift from the actual values. Here we quantify this drift again with 10 circular scans on laser-printed paper. We compare the distance between the first and last images of the online mosaicking with that of the offline mosaicking. Over the ten measurements the mean error is 12.59 μm with a standard deviation 4.14, a minimum 3.80 and a maximum 19.84. We observe that the mean and maximum errors are again far less than the image size. It should be noted that in this paper we use the online mosaicking algorithm only for distance measurement to determine the loading-distance that will shortly be explained.

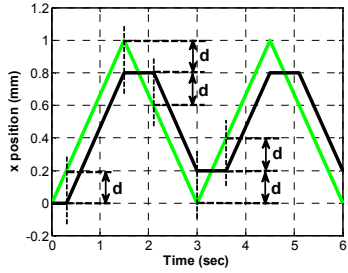

Fig. 3. Hypothetical soft-tissue behavior under surface contact with a scanning probe. Light (green) lines: the trajectory of the probe with respect to the global reference frame; dark (black) lines: the trajectory of the probe with respect to the soft tissue surface.

Based on the results in this section we can conclude that the offline mosaicking algorithm generates mosaics acceptably close to the ground truth images, the robot and online position measurements are precise enough considering the size of a single shot image.

## IV. LOAD-SLIP PHENOMENON ON SOFT TISSUE

It is demonstrated in finger slip studies [27, 28, 29] that when a finger is dragged on a solid surface on a straight line, the soft tissue of the finger experiences two successive phases: loading and slipping. In the loading phase the central part of the tissue remains stuck to the surface. Throughout loading, the drag force applied on the finger remains less than the static friction introduced by the sticking regions; therefore the finger does not slip on the surface. The soft tissue deforms under the impact of the friction force and is loaded with stress. When the central part also slips, the finger enters into the slipping phase. In this phase the drag force is equal to the kinetic friction. There is a stationary stress on the soft tissue and it remains throughout the movement in the slipping phase.

The behavior of a soft tissue being scanned with a solid probe is very similar. In this case the probe is dragged on the soft tissue. The stress is accumulated on the surface being scanned, rather than on the moving part. When the probe starts being dragged, the contact surface remains stationary with respect to the probe, because it sticks to the probe and moves with it. This is the loading phase of the scan. In this phase the static friction force stretches the contact surface and the tissue is loaded with stress. When the dragging force overcomes the friction, the probe starts moving with respect to the tissue surface and the movement enters into the slipping phase. The soft tissue is constantly loaded and unloaded with the movement of the probe. Unlike the case of the finger, the stress propagates on the soft tissue throughout the slipping phase.

When we apply the idea of load-slip phenomenon to the case of a line scan, the ideal behavior would look like in the hypothetical graph in Fig. 3. In this graph the light (green) lines correspond to the probe trajectory and the dark (black) lines correspond to the image trajectory. At the very start, the tissue sticks to the probe. Therefore the probe moves but the image collected by the probe (and visualized on the monitor) does not move. During this phase the tissue is loaded. When the probe reaches a distance of $d$ the tissue is fully loaded and enters into the slipping phase. From this point on the image follows the probe with the same speed, with a position lag $d$, up to the point that the latter stops. At 1.5 sec, the probe stops and starts moving in the opposite direction ($-x$). With the reverse movement of the probe the image stops moving, because the tissue remains stuck to the probe. The tissue starts unloading the stress. After the probe moves a distance $d$, the stress is totally unloaded. At this moment the probe and the image trajectories have the same position (0.8 mm). The probe goes further in the $-x$ direction and starts loading the tissue in this direction. Loading continues till the moment that the probe traverses the distance $d$. Afterwards the movement enters into the slip phase and the image again starts moving following the probe.

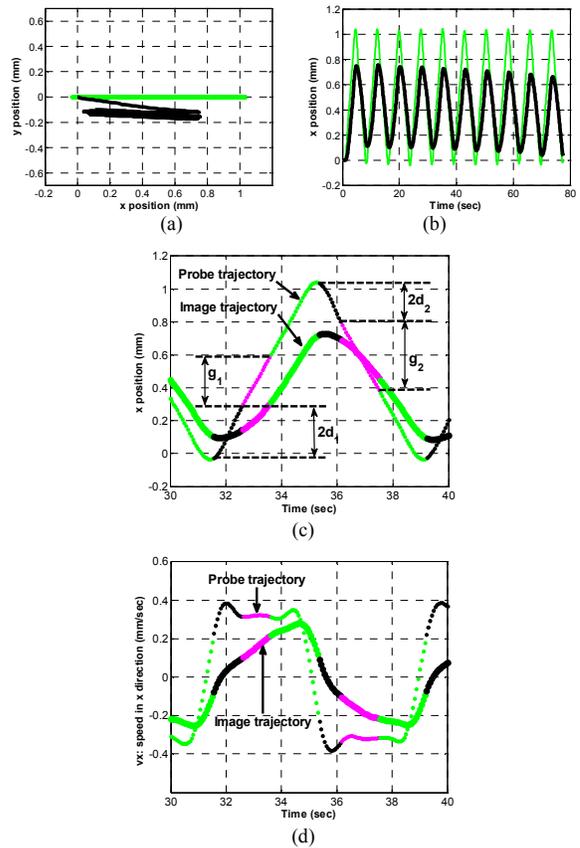

Fig. 4. Experimental results of 1 mm line scan on beef liver. (a) and (b): Light (green) line is the probe trajectory; dark (black) is the image trajectory. (c) The position profile in $x$-direction; (d) velocity profile in $x$-direction. In (c) and (d): the thin lines represent the probe movement; the thick lines represent the image movement; light (green) parts represent the slipping phases; dark (black) parts represent the loading-unload phases; gray (pink) parts represent the load+slip phases.

In Fig. 4 we demonstrate the actual result of a 10 times repeated forward-backward line scan in $x$-direction with 1 mm distance and 0.3 mm/sec speed of the probe with respect to the global reference frame. We observe that the position graphs in $x$-direction (Fig. 4(c)) are similar to the hypothetical one described above. For the purpose of closer examination of the speed relations we identify three different phases and designate them by three different colors in Fig 4 (c, d).

1) **Slipping phase:** *The difference between the probe and image speeds is less than 0.1 mm/sec.* In this phase the image trajectory closely follows the probe. The drag force is equal to the kinetic friction force between the probe and the tissue. This phase is designated by the light colored (green) lines on Fig. 4 (c, d).

2) **Loading-Unloading phase:** *The difference between the probe and image speeds is more than 0.1 mm/sec but the image speed is less than 0.1 mm/sec.* In this phase the probe is moving but the image is almost stationary. This corresponds to the regions where the tissue unloads and loads, being stuck to the probe. The drag force remains less than the maximum static friction force. This phase is designated by the dark (black) lines in Fig 4 (c, d). The distances corresponding to this phase on the probe trajectory are designated by $2d_i$, because this distance can be expected to correspond to the double of the loading-distance, $d$, shown for the hypothetical case.

3) **Load+slip phase:** *The difference between the probe and image speeds is more than 0.1 mm/sec but the image speed is more than 0.1 mm/sec.* In this phase both the probe and the image are moving but the speed of the image is less than that of the probe. This means that the probe is partially slipping on the tissue and partially loading the tissue. That is a phenomenon not noted in the finger slip studies. This phase is designated by gray (pink) lines in Fig. 4 (c, d). The distances corresponding to the load+slip phases in the probe trajectory are designated by $g_i$. The drag force is equal to the kinetic friction force but both of them steadily increase up to the point that the tissue is fully loaded. It can be expected that the kinetic friction force increases with increasing speed of slip from the beginning to the end of this phase. This is in agreement with the observation in [27] that the friction force slightly increases with increasing speed of slip.

If the actual measurements in Fig. 4 corresponded exactly to the ideal case in Fig. 3, we could designate the loading distance simply by taking the average of $d_1$ and $d_2$ values. However, the load+slip phenomenon observed in the actual case necessitates taking into account the partial loading in the load+slip phase. In this phase, the image speed starts from a low value close to zero and linearly increases to a value close to the robot speed. Due to this linearity, we can assume that half of the distance covered in the load+slip phase corresponds to loading (zero image speed), and the other half corresponds to slipping (image speed equal to the robot speed). Therefore we can assume that the load+slip phase contributes to the loading-distance with an amount of average $g_i/2$, half of the distance covered by the probe. Bringing together the contributions of the loading-unloading and load+slip phases, the loading-distance can be calculated as in (1).

$$d = \frac{d_1 + d_2}{2} + \frac{1}{2} \cdot \frac{g_1 + g_2}{2} \quad (1)$$

In the middle column of Table I, we present the loading-distance values calculated using (1). The experiments leading to the values in Table I are line scans like the one in Fig. 4(a, b), but with different speed and distance values, on a sample beef liver tissue. We name using (1) as the method of *phase designation*. The average loading-distance for this tissue is determined to be 0.191 mm with a standard deviation 0.047 mm across all constant speed and constant distance experiments.

Phase designation is exact with respect to our model of speed relations but difficult to implement because the speed measurement is noisy. It needs filtering and careful monitoring of the results to ensure that there is no discontinuity in the separately distinguished regions. A simpler method can be to use peak to peak distances of the position signals to approximate $d$. In Fig. 4(b) we observe that the difference between the peak to peak distances of the two trajectories is almost constant throughout the scan. This difference corresponds to the region traversed by the probe but not by the image. Therefore, it is closely related to the loading-distance. Considering both ends of the scan line, we can approximate the loading-distance by half of the difference between the peak to peak distances. This difference can easily be extracted from a period of the line scan. We name this approach as the method of *difference of peak to peak* and present the corresponding values in the third column of Table I. The average loading-distance calculated with this method is 0.205 mm with a standard deviation 0.051 mm. This value differs from the one calculated by phase designation by less than 5%. Therefore, the method of *difference of peak to peak* might be used instead of the *phase designation*.

TABLE I
LOADING-DISTANCE VALUES CALCULATED WITH TWO DIFFERENT METHODS FOR A SAMPLE BEEF LIVER TISSUE

| | Constant Distance Line Scan (1 mm) | |
|---|---|---|
| | Loading-distance | |
| Scan speed (mm/sec) | Phase designation (mm) | Difference of peak to peak (mm) |
| 0.20 | 0.179 | 0.187 |
| 0.25 | 0.163 | 0.190 |
| 0.30 | 0.171 | 0.198 |
| 0.35 | 0.165 | 0.183 |
| 0.40 | 0.197 | 0.188 |
| 0.45 | 0.205 | 0.179 |
| 0.50 | 0.225 | 0.173 |
| Mean | **0.186** | **0.185** |
| Std. dev. | 0.023 | 0.008 |
| | Constant Speed Line Scan (0.30 mm/sec) | |
| | Loading-distance | |
| Scan distance (mm) | Phase designation (mm) | Difference of peak to peak (mm) |
| 0.50 | 0.136 | 0.140 |
| 0.75 | 0.154 | 0.188 |
| 1.00 | 0.172 | 0.192 |
| 1.25 | 0.190 | 0.235 |
| 1.50 | 0.193 | 0.244 |
| 1.75 | 0.334 | 0.356 |
| 2.00 | 0.190 | 0.213 |
| Mean | **0.196** | **0.224** |
| Std. dev. | 0.065 | 0.068 |
| | Overall | |
| Mean | **0.191** | **0.205** |
| Std. dev. | 0.047 | 0.051 |

*A. Protocol for Loading-distance Measurement*

In the following we propose a protocol for determination of the loading-distance based on the method of *difference of peak to peak*. This protocol consists of a single forward-backward line scan lasting approximately 10 seconds.

*Protocol Steps:*

1) Determine the speed of scan,

2) Make the robot drag the probe linearly with a distance $d_r/2$; name the direction as the forward direction (forward direction is positive, backward direction is negative; $d_r$ can typically be chosen 0.5 mm),

TABLE II
STATISTICS FOR REPEATABILITY MEASUREMENTS OF LOADING-DISTANCE ON PAPER AND CHICKEN BREAST

| | Loading-distance (μm) | | | |
|---|---|---|---|---|
| | Paper | Chicken breast (locations) | | |
| | | I | II | III |
| Mean | 3.53 | 207 | 340 | 264 |
| Std. dev. | 3.75 | 10 | 7 | 4 |
| Max. | 0.65 | 191 | 332 | 259 |
| Min. | 14.22 | 220 | 350 | 273 |

3) Start recording the position corresponding to the image motion,
4) Make the robot drag the probe linearly with a distance $d_r/2$ in the forward direction and $d_r$ in the backward direction,
5) Stop recording the image motion,
6) Determine the minimum ($p_{min}$) and maximum ($p_{max}$) position values for the image motion along the scan line,
7) The loading-distance is given by (2).

$$d = \frac{d_r - (p_{max} - p_{min})}{2} \quad (2)$$

This protocol simply takes half of the difference between the distances covered by the probe and the image in one shot of the line scan. As it is easy to code and a fast procedure, it can be adapted to an automated scan.

### B. Repeatability of the Loading-Distance Measurement

In this section we test whether the loading-distance measurement with the above protocol provides repeatable results when performed on the same area of a given piece of tissue, with the same distance ($d_r$=1mm) and speed (0.3 mm/sec) line scan, under the same pressure and hydration conditions as explained in Section II.

The loading distance is affected by the surface and bottom structures of the tissue. Therefore we can expect the loading distance to slightly change across different locations of the same piece and across different pieces. However, considering a single location on the same tissue the loading distance is expected to remain the same across different measurements. If this is the case, we can use the same loading distance value for a scan on a given location.

We performed 10 measurements on the grid patterned paper, 10 measurements on each of two different parts of the same chicken breast tissue (location-I and -II), and 9 measurements on the same part of another chicken breast tissue (location III). The mean, standard deviation, maximum, and minimum of these sets of measurements are given in Table II. As expected, the measured loading distance is almost null on the rigid paper, whereas it is between 200 and 350 μm on the deformable tissues. Several estimations of $d$ on a same location give consistent values with low standard deviation. The maximum error in these measurements is on location-I, with 7.7% deviation from the mean value. Therefore we can state that the loading-distance measurement by the protocol is repeatable with an error margin in the order of 10%.

## V. TESTS ON BEEF LIVER AND CHICKEN BREAST

In this section we perform loading-distance measurements using the protocol with various scans on different locations of the same beef liver and chicken breast tissues. The locations differ with a distance of around 1 cm. We again perform the same indentation and hydration measures as explained in Section II. With these measurements we investigate the dependency of the loading-distance on the scan speed, scan distance, scan location and tissue type. The resulting loading-distance values are given in Table III and plotted in Fig. 5. The letters A, B, and C in Table III designate the three different locations on the same tissue.

TABLE III
LOADING-DISTANCE VALUES FOR BEEF LIVER AND CHICKEN BREAST, ON DIFFERENT LOCATIONS (A, B, AND C) ON THE SAME TISSUE

| | *Constant Distance* Line Scan (1 mm) | | | | | |
|---|---|---|---|---|---|---|
| | Loading-distance | | | | | |
| Speed | Beef liver (locations) | | | Chicken breast (locations) | | |
| (mm/sec) | A | B | C | A | B | C |
| 0.20 | 0.157 | 0.108 | 0.156 | 0.161 | 0.174 | 0.219 |
| 0.25 | 0.117 | 0.099 | 0.157 | 0.141 | 0.157 | 0.197 |
| 0.30 | 0.134 | 0.098 | 0.156 | 0.169 | 0.138 | 0.199 |
| 0.35 | 0.153 | 0.132 | 0.162 | 0.182 | 0.144 | 0.151 |
| 0.40 | 0.150 | 0.126 | 0.155 | 0.190 | 0.141 | 0.155 |
| 0.45 | 0.124 | 0.107 | 0.144 | 0.149 | 0.158 | 0.207 |
| 0.50 | 0.117 | 0.075 | 0.141 | 0.160 | 0.136 | 0.218 |
| Mean | 0.136 | 0.107 | 0.153 | 0.165 | 0.150 | 0.192 |
| Std. dev. | 0.017 | 0.019 | 0.007 | 0.017 | 0.013 | 0.028 |
| *Mean* | 0,132 | | | 0,169 | | |
| *Std. dev.* | 0,025 | | | 0,027 | | |
| | *Constant Speed* Line Scan (0.3 mm/sec) | | | | | |
| | Loading-distance | | | | | |
| Distance | Beef liver (locations) | | | Chicken breast | | |
| (mm) | A | B | C | | | |
| 0.50 | 0.062 | 0.113 | 0.083 | 0.114 | | |
| 0.75 | 0.094 | 0.131 | 0.108 | 0.153 | | |
| 1.00 | 0.097 | 0.137 | 0.123 | 0.110 | | |
| 1.25 | 0.087 | 0.149 | 0.123 | 0.157 | | |
| 1.50 | 0.099 | 0.175 | 0.128 | 0.184 | | |
| 1.75 | 0.155 | 0.213 | 0.162 | 0.168 | | |
| 2.00 | 0.181 | 0.221 | 0.192 | 0.171 | | |
| Mean | 0.111 | 0.162 | 0.131 | 0.151 | | |
| Std. dev. | 0.042 | 0.042 | 0.036 | 0.028 | | |
| *Mean* | 0.135 | | | 0.151 | | |
| *Std. dev.* | 0.044 | | | 0.028 | | |
| | *Overall* | | | | | |
| **Mean** | **0.133** | | | **0.164** | | |
| Std. dev | 0.035 | | | 0.028 | | |

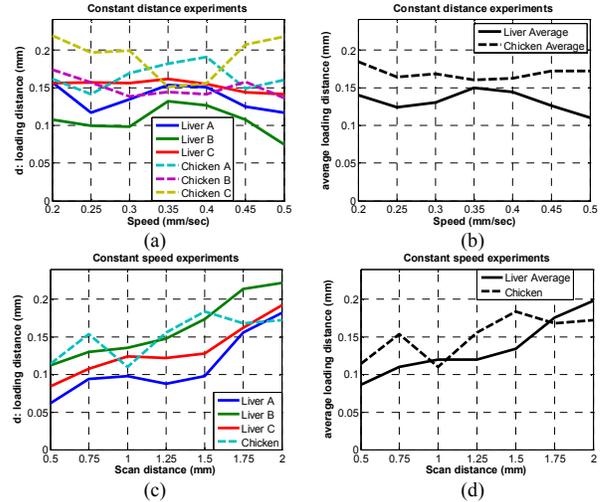

Fig. 5. Loading-distance values for various line scans on beef liver and chicken breast. (a) Constant distance (1 mm) line scans with different speed at three different locations (A, B, C) on each tissue; (b) average values of loading-distance for constant distance (1 mm) line scans with different speed; (c) constant speed (0.3 mm/sec) line scans with different scan distance at three different locations on beef liver (A, B, C) and single location on chicken breast; (d) average values of loading-distance for constant speed (0.3 mm/sec) line scans with different scan distance.

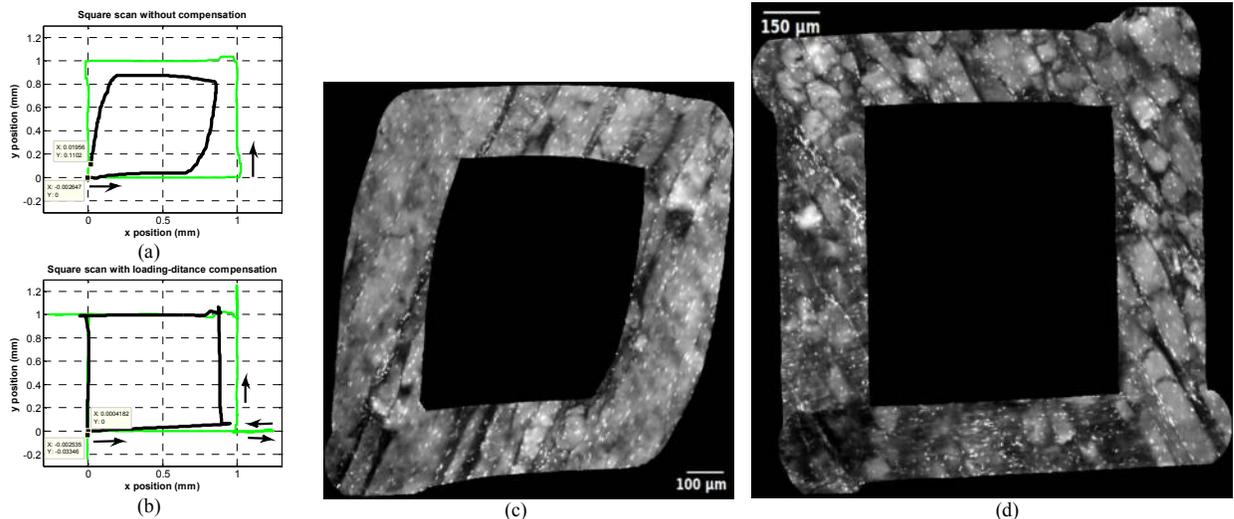

Fig. 6. Square scan results on chicken breast tissue. a) Square scan without compensation; (b) square scan with compensation with a loading-distance 0.25 mm. Light (green) lines: the probe trajectory; dark (black) lines: image trajectory. (c) Image-mosaic corresponding to the square scan without compensation; (d) image-mosaic corresponding to the square scan with compensation.

In [27], it is demonstrated that, in the case of finger, the friction force in the slipping phase slightly increases with increasing speed. This observation made us expect that the loading-distance would increase with increasing speed. However, Fig. 5(a) shows that it remains almost constant across varying speeds of scan, with an average 0.132 mm and standard deviation 0.025 for the beef liver and with an average 0.169 mm and standard deviation 0.027 for the chicken breast (Table III). Fig. 5(b) shows the average values separately for the beef liver and chicken breast. We observe that the average loading-distance for chicken breast is consistently larger than that of the beef liver across all speed values. This is expected since the chicken breast has a stickier surface compared to the beef liver.

Fig. 5(c) shows the dependency of the loading-distance on the scan distance with a constant scan speed 0.3 mm/sec. We observe that the loading-distance slightly increases with increasing scan distance. The average values for varying scan distance are 0.135 mm with a standard deviation 0.044 for beef liver and 0.151 mm with a standard deviation 0.028 for chicken breast. These values differ from the average values with varying scan speed (0.132 and 0.169 mm, respectively) by less than 5%. The closeness of the averages across different speeds and distances can be considered as an indication that loading-distance might be used to characterize a given pieces of tissue under specified scan conditions. Without verification of this idea we note here that, when we consider all speed and distance experiments, the average loading-distance is found to be 0.133 mm with a standard deviation 0.035 for the piece of beef liver and 0.164 mm with a standard deviation 0.028 for the chicken breast.

The average loading-distance measured at different locations of the same piece of tissue slightly differs. The reasons for this might be that, 1) there are slight thickness differences on the bottom part of the different locations, 2) there are differences on the surface structure, 3) and we do not maintain the same amount of pressure due to our error margin of 100 μm indentation depth. It should also be noticed that the average loading-distance for the beef liver used in the experiments of Table III (0.133 mm) differs from that of the beef liver used in the experiments of Table I (0.205 mm). The reason for this is probably that the bottom structures are different in size and shape. This observation points out that the loading-distance we measure might differ across different pieces of the same type of tissue depending on the size and shape.

Considering measurements on single location, the variations in Table III are larger than those observed in the results presented in Table II. This should be considered as the impact of the changing speed and scan distance. This is an indication that the best practice would be to perform the loading-distance measurement with the speed and line distance that applies to the intended scan. When the tissue is changed, the loading-distance should be determined anew. The results also suggest performing a new measurement when the scan location is changed on the same piece. All these imply that the protocol we present is best to be used as a calibration procedure prior to each scan.

## VI. APPLICATION TO RASTER SCAN

In this section we present the application of loading-distance compensation to raster scans. We first explain the compensation action with the square scan in Fig. 6 on a chicken breast tissue. Then we demonstrate the application to full raster scans in Fig. 7 and Fig. 8 on chicken breast and beef liver tissues respectively.

If the knowledge of the loading-distance is not used and the robot is commanded to perform a square scan, the image trajectory will not be a proper square. This is illustrated in Fig. 6(a). The dark (black) line representing the image trajectory significantly deviates from the light (green) line representing the probe trajectory. The edges are not in the desired length 1 mm and the corners are not perpendicular. The end point of the image trajectory is quite far from the start: the distance is more than 110 μm in Fig. 6(a). The resulting mosaic is shown in Fig. 6(c).

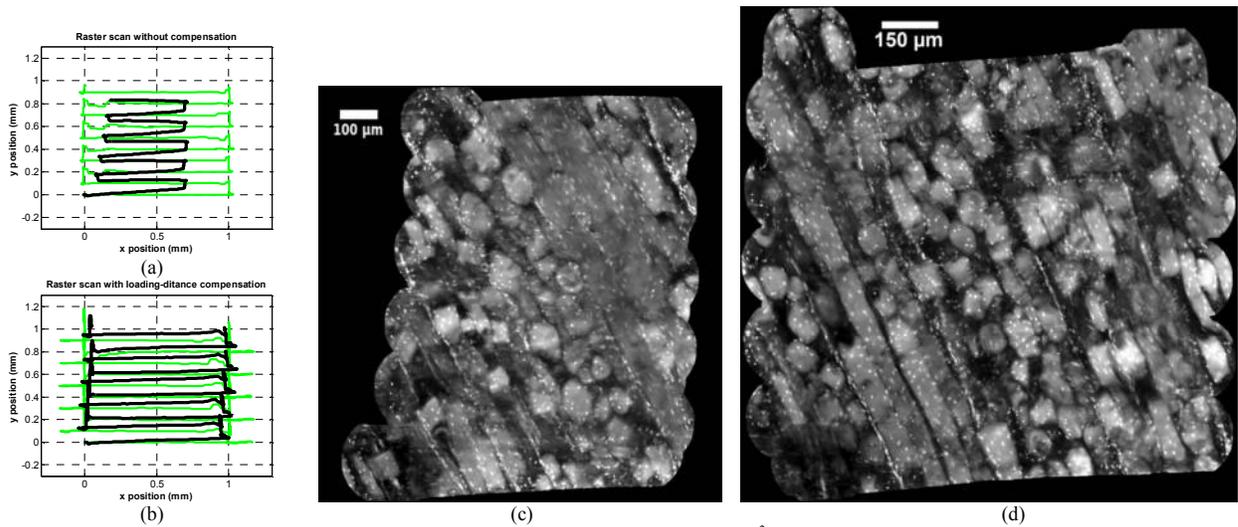

Fig. 7. Raster scan results on chicken breast corresponding to an area of approximately 1 mm$^2$ with 100 µm distance in between the scan lines. (a) The probe and image trajectories for the scan without compensation; (b) the robot and image trajectories for the scan with compensation; (c) the mosaic image for the scan without compensation; (d) the mosaic image for the scan with compensation. The loading distance for the scan in (b) and (d) is automatically determined by the integrated calibration protocol and realized as 165 µm

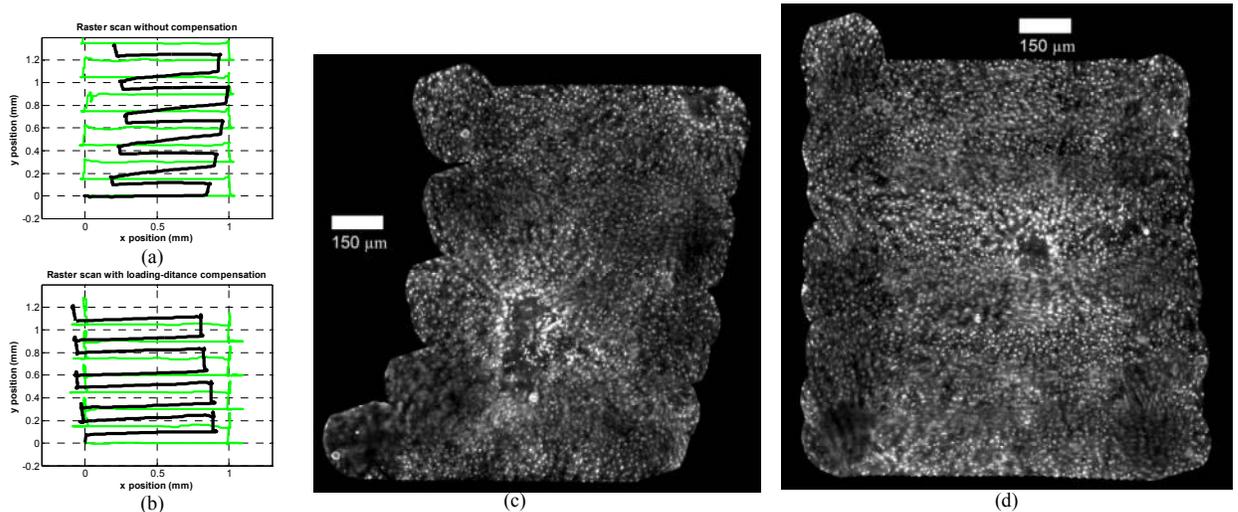

Fig. 8. Raster scan results on beef liver corresponding to an area of approximately 1 mm$^2$ with 150 µm distance in between the scan lines. (a) The probe and image trajectories for the scan without compensation; (b) the robot and image trajectories for the scan with compensation; (c) the mosaic image for the scan without compensation; (d) the mosaic image for the scan with compensation. The loading distance for the scan in (b) and (d) is automatically determined by the integrated calibration protocol as 61 µm; the robot realizes a loading-distance compensation of around 90 µm.

In order to achieve a better square with the image trajectory we modify the probe trajectory by a length of the loading-distance at the corners. This compensation action is indicated by the arrows at the corner in Fig. 6(c). The loading-distance, $d$=0.25 mm, is determined using the protocol prior to the scan. The robot compensation is sometimes slightly larger than the measured loading-distance due to the 50 µm error margin with our control. When the probe reaches a corner, it translates further in the current direction by a loading-distance and then turns back to the target corner by translation the same amount in the reverse direction. At this moment the tissue is unloaded. Then, the probe starts translation through the next perpendicular edge. Since the tissue is unloaded before starting each new edge, the image trajectory closely follows the probe trajectory. The improvement is clearly observed in Fig. 6(b). The probe trajectory closely follows the intended square. The length of the edges are much closer to the intended 1 mm and the corners are perpendicular. The distance between the start and end points is less than 35 µm. The corresponding mosaic, in Fig. 6(d) is a much better square image compared to the one without the correction.

In the raster scan experiments of Fig. 7 and Fig 8, we aim to cover a square area with edges 1 mm long on beef liver and chicken breast tissues, respectively. The distances between the scan lines are 150 µm in Fig. 7 and 100 µm in Fig. 8. These are close enough to have an overlap between the images of the upper and lower lines of the raster scan with the 240×200 µm$^2$ field of view. When the robot is commanded to follow the scan lines without any compensation, the image trajectory deviates from the probe trajectory (Fig. 7(a) and Fig. 8(a)). The covered area on the tissue is much less than the intended and the shape is not a square (Fig. 7(c) and Fig. 8(c)).

For the compensated raster scan experiments presented in Fig. 7(b) and Fig. 8(b), we integrate the programs for automatic loading-distance measurement and for automatic scan. In this way the measurement and scan are performed with a single action. In other words the system first calibrates by measuring the loading-distance and then performs the scan with compensation. The overall scan last less than one minute (~10 seconds calibration plus ~40-50 seconds compensated raster scan, with 0.3 mm/sec scan speed). We apply the loading-distance compensation at all corners of the successive rectangular paths. It is observed in Fig. 7(b) and Fig. 8(b) that the resulting path consists of shapes that are close to rectangles. The long-side edges have almost the intended length; the corners have almost right angles; and the short-side edges are satisfactorily long. All these result in the mosaics in Fig. 7(d) and Fig. 8(d) that cover the intended square with straight and 1 mm long edges.

## VII. Conclusion

In this paper we present an approach to understand the soft tissue behavior in surface contact with a probe for microlaparoscopic scan. Our approach is inspired by the load-slip phenomenon observed in finger slip studies. We apply the idea of load-slip phenomenon to our soft tissue scan experiments and we develop the parameter of loading-distance. This parameter provides explanation for the deviation between the probe and image trajectories throughout the scan. We present various ex vivo loading-distance measurements with varying speed and distance, on beef liver and chicken breast tissues. Our results provide strong evidence that the loading-distance remains constant over the range of a practical scan area on soft tissue. We propose and implement a protocol to measure the loading-distance prior to an automated confocal microlaparoscopic scan. This protocol is simple to be used and programmed. It can be performed in around ten seconds prior to the actual scan and the measured loading-distance can be used throughout the scan. We demonstrate the effectiveness of using the loading-distance with raster scan experiments. For this purpose, we integrate the proposed protocol with a compensation action to the automated raster scan procedure. The system automatically measures the loading-distance and adapts the compensation to the measured value. The results demonstrate the improvement in the mosaic image.

The loading-distance slightly varies in different locations of the same piece of tissue. We believe that the slight difference in the pressure is the cause of changes on the same piece of tissue. We observed in our experiments that the loading distance changes with pressure and with the level of lubrication. When the robot had more indentation (larger pressure) the loading distance clearly increased. When the tissue was left aside to dry, it became very sticky (especially the chicken breast) and the loading distance again increased. The loading-distance significantly varies across different pieces of the same type of tissue. The cause of this latter variation is expected to be the different shape and thickness of the pieces used. Despite these observations, in this paper we did not aim at a quantification of the impacts on loading-distance. In this paper, demonstrating the practical usage of this parameter was prior to investigating its dependence on changing physical conditions. The latter remains as a future work.

We consider that the change of the loading-distance across different types of tissue (beef liver and chicken breast) is natural. This is because, first, the surface structures are different, second, the same indentation depth causes different pressure due to the differences in the tissue structure. The question is whether the loading-distance can be used as a characterizing parameter for the tissue structures, under specified conditions of pressure, hydration, and tissue shape. Here we only point out to this possibility and we leave it as a future work.

For application of our method, it is not necessary that the loading-distance is a characterizing measure. What we need is that it remains constant within the limited area of a typical scan (~1 mm$^2$) and we show that this is indeed the case in practice. Therefore, we make the loading-distance measurement at the start of the scan and use the same value throughout the scan. In cases where a very large area scan is needed, an approach would be to perform loading-distance measurements in different phases of the scan. Our approach can easily be adapted for that. In this way our approach can be used for any large area scan without being restricted to have the same length loading-distance everywhere.

We are further researching the systematic ways of using the knowledge of loading-distance to compensate any sort of movement on soft tissue, especially those following curved trajectories.

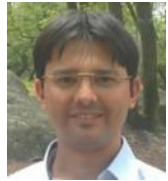

Mustapha Suphi Erden received the B.S., M.S., and Ph.D. degrees in Electrical and Electronics Engineering from Middle East Technical University, Ankara, in 1999, 2001, and 2006. Between 1999 and 2006, he was a Research Assistant in the same department. From 2007 to 2012 he has been a postdoctoral researcher, successively in Delft University of Technology, the Netherlands; in Ecole Nationale Supérieure de Techniques Avancées-ParisTech, France; in Univ. Pierre & Marie Curie – Paris 6. In 2012 he received the European Union Marie Curie Intra-European Fellowship. Since September 2012, he is with the Learning Algorithms and Systems Laboratory, Ecole Polytechnique Fédérale de Lausanne, Switzerland, with this fellowship. His research interests include human-robot interaction, assistive robotics, skill assistance, mechatronics design, medical robotics, walking robots, and machine learning.

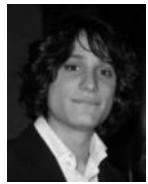

Benoît Rosa obtained his Engineer's degree in Electrical and Electronical Engineering at Ecole Centrale de Paris, France, in 2009. Since 2010, he is working toward the Ph.D. degree at Institut des Systèmes Intelligents et de Robotique, Univ. Pierre & Marie Curie – Paris 6. He is also Teaching Assistant at Univ. Pierre & Marie Curie – Paris 6. His research interests include medical and surgical robotics, biomedical engineering, microactuation and visual servoing.

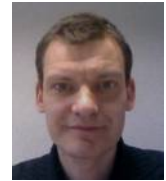

Jérôme Szewczyk received a M.S. degree in Mechanical Engineering from Univ. of Compiègne (France) in 1994 and a Ph.D. degree in robotics at Univ. Pierre & Marie Curie – Paris 6 in 1998. Between 2000 and 2010, he was Assistant Professor at Univ. of Versailles (France). Since 2011 he is Associate Professor at Univ. Pierre & Marie Curie – Paris 6. His research activities at Institut des Systèmes Intelligents et de Robotique concern the design of sensors and actuators for meso-robotics based on smart materials and the optimal design of robotic structures for surgical applications.

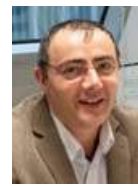

Guillaume Morel obtained his Ph.D. degree in Control Engineering in Univ. Pierre & Marie Curie – Paris 6, France in 1994 and went for a postdoctoral period at the Massachusetts Institute of Technology, USA, in 1995-1996. Back in France, he was successively a research engineer for the French Company of Electricity and an Assistant Professor at Univ. of Strasbourg, France (1997–2001). He is now Professor in Robotics at Univ. Pierre & Marie Curie – Paris 6. His research interests have always been in the sensor based control of robots, with a particular focus on force feedback control and visual servoing. Since 2000, his research target applications are assistance for surgery and, more recently, rehabilitation systems. Today, within the Institute of Intelligent Systems and Robotics, he leads the research team AGATHE (Assistance to Gesture with Application to THErapy), which develops the concept of comanipulation, where a robot and a human user share the control of a same object for the realization of a manipulation task.